\documentclass{llncs}

\usepackage[a-2u]{pdfx}
\usepackage[utf8]{inputenc}
\usepackage{lmodern}
\usepackage[T1]{fontenc}
\usepackage{xcolor}

\usepackage{booktabs}
\usepackage{multirow}
\usepackage{graphicx}

\usepackage{microtype}
\usepackage{listings, multicol}
\definecolor{commentcolor}{rgb}{0.5,0.5,0.5}
\definecolor{darkgreen}{rgb}{0.09, 0.45, 0.27}
\definecolor{bgcolor}{rgb}{0.99,0.99,0.99}
\lstdefinestyle{ensembles}{
    morekeywords={situation,constraints,constraint,allow,utility,notify,rules,ensembles,notified,oneOf,unionOf, subsetOfComponents,tasks},
    basicstyle=\linespread{0.8}\sffamily\scriptsize,        % the size of the fonts that are used for the code
    breakatwhitespace=true,           % sets if automatic breaks should only happen at whitespace
    showstringspaces=false,
    columns=fullflexible,
    tabsize=2,%
    breaklines=true,
    captionpos=b,
    numbers=left,
    numbersep=5pt,
    numberstyle=\linespread{0.8}\sffamily\tiny\lsstyle,
    xleftmargin=0.05\columnwidth,
    commentstyle=\color{commentcolor},
    keywordstyle=\color{blue},
    stringstyle=\color{darkgreen},
}
\newcommand{\longlstinline}[1]{{\small\textsf{#1}}}
\usepackage{subcaption}
\usepackage{footmisc}
\usepackage{paralist}

\usepackage{fancyhdr}
\fancypagestyle{plain}{%
\fancyhead{}
\fancyfoot{}

\fancyfoot[L]{\scriptsize This is the authors' version of the paper: \textit{T. Bureš, I. Gerostathopoulos, P. Hnětynka, J. Pacovský. Forming Ensembles at Runtime: A Machine Learning Approach, in Proceedings of ISOLA 2020, Rhodes, Greece, 2020}.\\ The final authenticated publication is available online at \url{https://doi.org/10.1007/978-3-030-61470-6\_26}}
}
\fancypagestyle{empty}{%
\fancyhead{}
\fancyfoot{}

\fancyfoot[L]{\scriptsize This is the authors' version of the paper: \textit{T. Bureš, I. Gerostathopoulos, P. Hnětynka, J. Pacovský. Forming Ensembles at Runtime: A Machine Learning Approach, in Proceedings of ISOLA 2020, Rhodes, Greece, 2020}.\\ The final authenticated publication is available online at \url{https://doi.org/10.1007/978-3-030-61470-6\_26}}
}

\begin{document}
\clearpage
\pagestyle{plain}
\title{Forming Ensembles at Runtime: A Machine Learning Approach}
\author{
Tomáš Bureš\inst{1},
Ilias Gerostathopoulos\inst{1,2},
Petr Hnětynka\inst{1},
Jan Pacovský\inst{1}
}
\institute{Charles University, Czech Republic\\ \email{\{bures,hnetynka,pacovsky\}@d3s.mff.cuni.cz}
\and
Vrije Universiteit Amsterdam, Netherlands\\ \email{i.g.gerostathopoulos@vu.nl}
}

\maketitle

\begin{abstract}
Smart system applications (SSAs) built on top of cyber-physical and socio-technical systems are increasingly composed of components that can work both autonomously and by cooperating with each other. Cooperating robots, fleets of cars and fleets of drones, emergency coordination systems are examples of SSAs. One approach to enable cooperation of SSAs is to form dynamic cooperation groups—ensembles—between components at runtime. Ensembles can be formed based on predefined rules that determine which components should be part of an ensemble based on their current state and the state of the environment (e.g., ``group together 3 robots that are closer to the obstacle, their battery is sufficient and they would not be better used in another ensemble''). This is a computationally hard problem since all components are potential members of all possible ensembles at runtime. In our experience working with ensembles in several case studies the past years, using constraint programming to decide which ensembles should be formed does not scale for more than a limited number of components and ensembles. Also, the strict formulation in terms of hard/soft constraints does not easily permit for runtime self-adaptation via learning. This poses a serious limitation to the use of ensembles in large-scale and partially uncertain SSAs. To tackle this problem, in this paper we propose to recast the ensemble formation problem as a classification problem and use machine learning to efficiently form ensembles at scale. 
\end{abstract}


\section{Introduction}
\label{sec:introduction}
Smart system applications (SSAs) are cyber-physical and socio-technical systems that comprise a number of components that cooperate towards a common goal. These systems are increasingly popular and include wide range of applications spanning from smart building management (coordinating heat, ventilation, air conditioning with physical access control, etc.), smart traffic, emergency response systems, up to smart farming or smart underwater exploration. 

The cooperation among components is a key feature of these systems. The cooperation is typically governed by certain application-specific collaboration rules. For instance, in the smart farming domain, the drones monitoring the crop on fields may coordinate to keep three drones in the air patrolling the surveyed area while the rest of the drones recharges or stays idle on the ground. This naturally leads to describing the coordination in these systems using a set of constraints (both hard constraints and soft constraints) that govern which components should cooperate at a particular point in time and what their task is. For instance, such constraints would assign the three drones the task to patrol the fields. The selection would be based on the battery level of the drone and its distance to the patrolled area. Formally speaking, the correct and optimal operation of such a system is then reduced to the problem of finding the assignment of components to collaboration groups that satisfy the constraints and optimize the soft-optimization rules.

In our work we use the architectural concept of autonomic component ensemble to model the collaboration group. The ensemble defines its potential members and stipulates the set of hard constraints and optimization rules that govern which components are eventually selected as members. As multiple ensembles can co-exist at the same time, there are naturally also constraints across the ensembles---for enforcing that the same component may be a member of only one of several ensembles of the same type. The problem of finding such assignment of components to ensembles (we term this \textit{ensemble resolution}) is inherently exponential and cannot be easily overcome even with state-of-the-art SMT or CSP solvers. 

The problem is even aggravated by the fact that the solution to the problem has to be found repeatedly---essentially whenever the state of the components or the environment changes. This is because the constraints controlling the membership of components in ensembles typically directly depend on the state of the components and the environment. As this state constantly changes, ensembles have to be continuously re-resolved at runtime. This puts hard practical limits on the time needed to complete the ensemble resolution to be in order of seconds (up to minutes), and consequently on the maximum size of the system (which, based on our experiments~\cite{hnetynka_using_2020}, depends on the complexity of ensembles often limited to a dozen or a few dozens of components). 
Longer waiting times means that the system cannot flexibly react to ever changing situations in its environment. 

In this position paper, we thus explore an approach to address the problem of ensemble resolution that does not require exponential time at runtime. In particular, we show how the problem of ensemble resolution can be recast to a classification problem. In our work we use both neural networks and decision trees as classifiers. After training the classifier offline, we can execute it quickly at runtime, thus significantly cutting down the time needed to resolve ensembles.

As we discuss further in the text, using the classifier conceptually changes the problem from crisp solutions that strictly follow the hard constraints to fuzzied solutions that do not necessarily have to strictly obey the hard constraints. 

As it turns out, if well designed, such a system with approximate solutions still works. This requires a bit more robust design that balances well the responsibilities in the system among autonomously operating components (meaning that the components themselves are responsible for ensuring their safe and reliable operation) and ensemble-level decisions that deal with high-level coordination of components. However, such a design is overall necessary to make the system more robust to uncertainty and to facilitate decentralization. 

In this paper we report on our initial experiments in this direction. To demonstrate and evaluate the idea, we adopt a use-case inspired by our work in a smart farming project~\cite{hnetynka_using_2020}. We use the agent-based simulator of the use-case scenario to draw indicative results pointing to the feasibility of our approach.

We describe our running example inspired by our work in the smart farming project~\cite{hnetynka_using_2020} in Section~\ref{sec:example}. Then we explain how to recast the problem of ensemble formation to a classification problem in Section~\ref{sec:methods}. Section~\ref{sec:evaluation} provides an initial evaluation of the feasibility of our prediction-based ensemble resolution. Finally, Section~\ref{sec:related-work} positions the work w.r.t related ones, and Section~\ref{sec:conclusion} concludes with a summary and outlook.

\section{Running example}
\label{sec:example}
\begin{figure}[h]
    \centering
    \includegraphics[width=0.7\columnwidth]{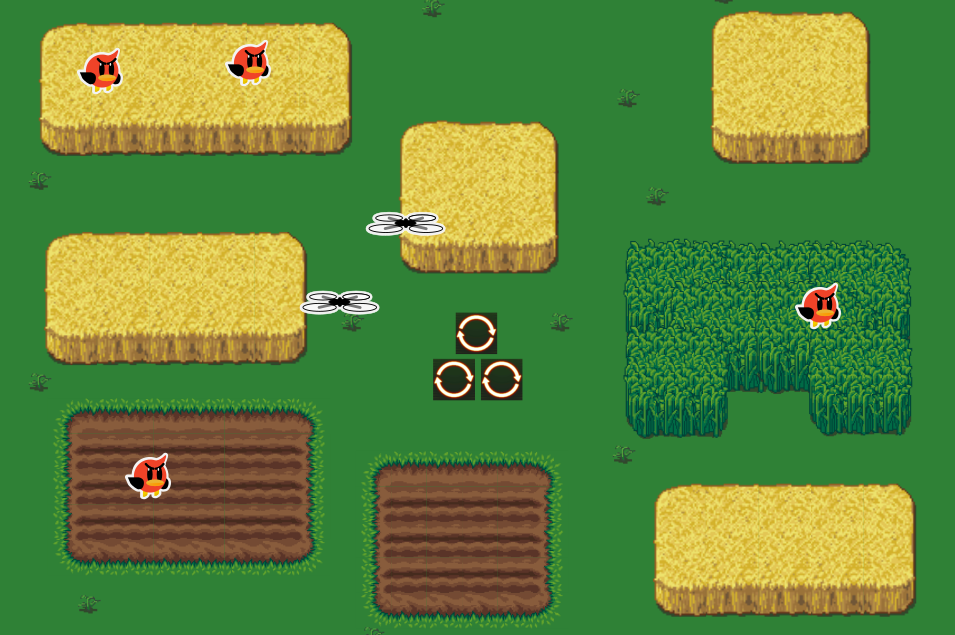}
    \caption{Example}
    \label{fig:example}
\end{figure}

As a motivational example, we use an actual scenario taken from our ECSEL JU project AFarCloud\footnote{\url{https://www.ecsel.eu/projects/afarcloud}}, which focuses on smart-farming and efficient usage of cyber-physical and cloud-systems in agriculture.
Figure~\ref{fig:example} shows a screenshot from our simulator developed to demonstrate the scenario.

In the scenario, there is a farm with several fields---the yellow ones represent fields with crop requiring a protection from birds (which can damage the crop) while the brown and dark-green ones require no protection.
The whole farm is continuously monitored by a fleet of autonomous drones.
The drones perform environment monitoring (humidity, temperature, etc.) and also detection of flocks of birds.
In case a flock is detected, the drones are used to scare the flock away from the crop fields to the birds-insensitive farm areas.
To be effective in scaring, the drones have to form a group (depending on the size of the flock). 
Additionally, the drones can operate for a limited time only (depending on the battery capacity) and need to periodically recharge in the charger (the rounded arrow blocks at the center), but which can charge only a limited number of drones at the same time.
Thus in order to be effective as a whole, the system has to balance between the number of drones monitoring the farm, scaring the birds, and charging themselves.
Plus, the system needs to select the most suitable drones for individual tasks (i.e. the closest ones, with a sufficient amount of energy, etc. depending on the task).

To model and run dynamic and autonomous systems (like this one), we use our approach based on autonomic ensembles~\cite{tomas_bures_language_2020}.
In this approach, entities of a system are modeled as components and cooperation among the components is modeled via ensembles, which are dynamic context-dependent (time and space bound) groups of components.
For simple and easy development and experimentation with ensembles, we have created a Scala-based internal domain-specific language (DSL) to specify components and ensembles.

Listing~\ref{lst:dsl} shows an excerpt of the specification of the example in DSL.
Both the component and ensemble types are modeled as classes, while the actual component and ensemble instances are instantiations of these classes (there can be a number of instances of a particular type).
In the example, there are four component types---DroneComponent, FieldComponent, ChargerComponent and FlockComponent (lines \ref{inlst:comp-start}--\ref{inlst:comp-end}).
A component state (called component knowledge) is modeled via the class fields.
The FieldComponent and FlockComponent are non-controllable components, i.e., they cannot be directly controlled by the system and their state is observed only.

\begin{figure}[h]
\centering
    \includegraphics[width=1\columnwidth]{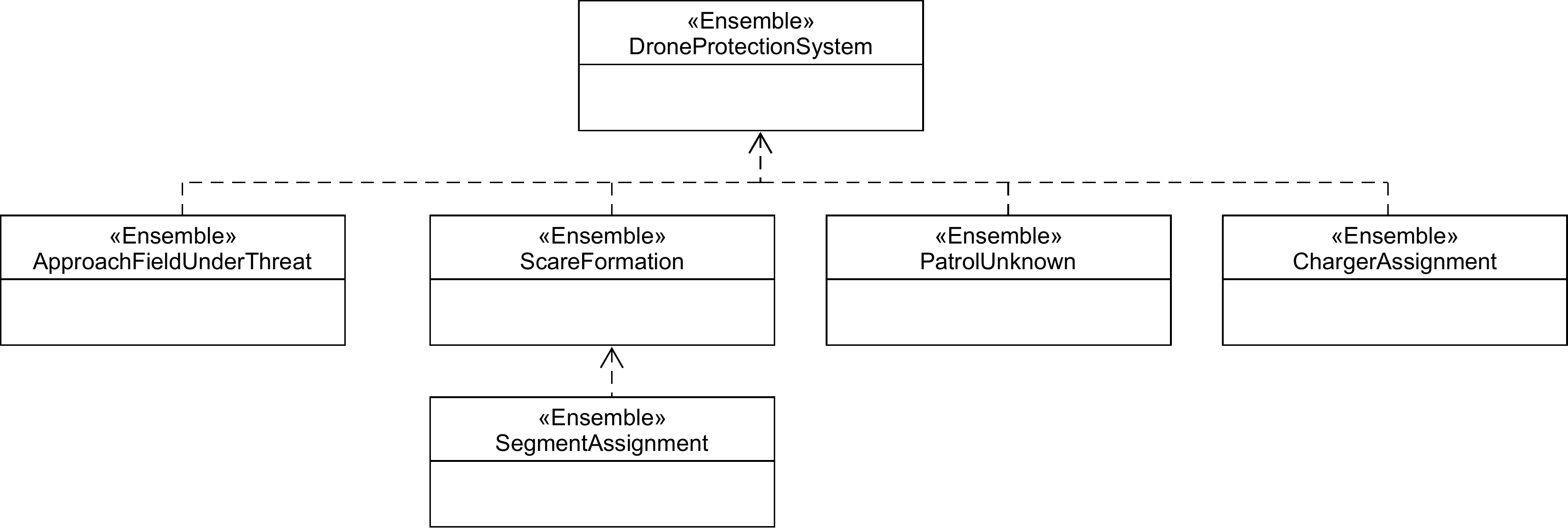}
    \caption{Ensembles hierarchy in the running example.}
    \label{fig:hierarchy}
\end{figure}

There are six ensemble types---the top level one and five nested ones (the structure is shown in Figure~\ref{fig:hierarchy}).
An ensemble is formed dynamically in order to execute a group-level behavior (e.g., scare the flock). Only the top-level ensemble instance (called the \textit{root ensemble}) exists for the whole lifetime of the system (it is instantiated at line~\ref{inlst:root}). 
The component instances grouped in the ensemble are not defined statically but via its membership condition, which is a conjunction of predicates over the component instances and their knowledge and which is continuously evaluated. For instance, in the \longlstinline{DroneProtectionSystem} ensemble, the lines \ref{inlst:members-start}--\ref{inlst:members-end} of Listing~\ref{lst:dsl} identify component instances in the system and categorize them by mode. These groups of components are then used in ensembles to declare potential members. The declaration of potential members is done via \longlstinline{oneOf} (e.g. line~\ref{inlst:oneOf}) and \longlstinline{subsetOfComponents} (e.g. line~\ref{inlst:subsetOf}). These two declarations differ only in cardinality constraints.

A particular component instance can be a member of several ensemble instances at the same time.
If additional conditions have to be hold, the \lstinline{constraint} construct can be used (e.g., at line \ref{inlst:constraint} in the \longlstinline{DroneProtectionSystem}).
The \lstinline{utility} construct (e.g., at line \ref{inlst:utility} in the \longlstinline{DroneProtectionSystem}) represents a soft condition, which is an optimization function for forming ensembles (i.e., in the case, there are several possibilities to choose component instances for the ensemble). Finally, the \lstinline{tasks} construct assigns responsibilities to component instances which are members of a particular ensemble instance.

As already mentioned, ensembles can be nested; members (i.e. component instances) of an ensemble are also members of a parent ensemble.
The meaning of nesting is that the root ensemble (the one without any parent) defines the overall goal of the system while the nested ones represent particular sub-goals.
The \longlstinline{DroneProtectionSystem} ensemble is the root and represents the whole system. 
The \longlstinline{ApproachFieldUnderThreat} ensemble is used to direct the necessary number of drones to the field with detected birds. 
When the necessary number of drones is above the field, an instance of the \longlstinline{ScareFormation} ensemble replaces the instance of the previous ensemble and tries to scare birds away from the field by moving drones above the field in a formation that equally spreads the drones over the field.
To simplify definition of assignment of the drones to positions over the affected field, the ensemble contains the sub-ensembles
\longlstinline{SegmentAssignment}.
The \longlstinline{PatrolUnknown} ensemble is used to guide a drone to a field that has unknown status (no information about birds).
And finally, the \longlstinline{ChargerAssignment} ensemble is instantiated per charger place and assigns a drone that has a low energy level to
the charger. 

The sub-ensembles are declared using \longlstinline{ensembles} (e.g. line~\ref{inlst:ensembles}) and \longlstinline{rules} (e.g. line~\ref{inlst:rules}). The former declares a potential ensemble instance (i.e. such that will be instantiated if constraints allow it), the latter declares a mandatory ensemble instance (i.e. such that has to exist if its parent ensemble instance gets instantiated).

\begin{lstlisting}[caption=Running example in DSL, label=lst:dsl, language=Scala, escapechar=|, breaklines=true, style=ensembles]
case class DroneComponent(|\label{inlst:comp-start}|
    id: String, mode: DroneMode.DroneMode, position: Position, energy: Double, 
    chargingInChargerId: Option[ChargerId], observedFields: Map[String, FieldObservation]
  ) extends Component {
  name(id)
}

case class FieldComponent(idx: Int, flocks: Map[String, FlockState]) extends Component {
  name(s"Field ${idx}")
  val center = FieldIdHelper.center(idx), val area = FieldIdHelper.area(idx)
  val isUnknown = false, val isUnderThreat = flocks.values.exists(flock => area.contains(flock.position))
  val requiredDroneCountForProtection = FieldIdHelper.protectingDroneCountRequired(idx) 
  val protectionCenters = FieldIdHelper.centers(idx, requiredDroneCountForProtection)
}

case class ChargerComponent(idx: Int, isFree: Boolean) extends Component {
  name(s"Charger ${idx}")
  val chargerId = ChargerId(idx), val position = chargerId.position
}

case class FlockComponent(position: Position) extends Component {
  name(s"Flock @ ${position}")
}|\label{inlst:comp-end}|

class Scenario(simulationState: SimulationState) extends /*....*/ {

  class DroneProtectionSystem extends Ensemble {
    name(s"Root ensemble of the protection system")

    val operationalDrones = allDrones.filter(drone => drone.mode != DroneMode.DEAD && drone.mode != DroneMode.CHARGING && drone.energy > Drone.chargingThreshold) |\label{inlst:members-start}|
    val dronesInNeedOfCharging = allDrones.filter(drone => drone.mode != DroneMode.DEAD && drone.mode != DroneMode.CHARGING && drone.energy < Drone.chargingThreshold)
    val fieldsWithUnknownStatus = allFields.filter(_.isUnknown)
    val fieldsUnderThreat=allFields.filter(_.isUnderThreat), val freeChargers=allChargers.filter(_.isFree)|\label{inlst:members-end}|

    class ApproachFieldUnderThreat(val field: FieldComponent) extends Ensemble {
      name(s"ApproachFieldUnderThreat ensemble for field ${field.idx}")

      val flocksInField = allFlocks.filter(x => field.area.contains(x.position))
      val dronesInField = operationalDrones.filter(x => field.area.contains(x.position))
      val droneCount = field.requiredDroneCountForProtection, val center = field.center
      val drones = subsetOfComponents(operationalDrones, _ <= droneCount)|\label{inlst:subsetOf}|

      utility {drones.sum(x=>if (field.area.contains(x.position)) 10 else dist2Utility(x.position, center))}

      tasks {
        if (flocksInField.isEmpty) {
          for (drone <- drones.selectedMembers) moveTask(drone, center)
        } else {
          val selectedDronesInFieldCount = drones.selectedMembers.count(x => field.area.contains(x.position))
          /* ... */
        }
      }
    }

    class ScareFormation(val field: FieldComponent) extends Ensemble {
      name(s"ScareFormation ensemble for field ${field.idx}")

      val dronesInField = operationalDrones.filter(x => field.area.contains(x.position))
      val droneCount = field.requiredDroneCountForProtection
      val segmentCenters = field.protectionCenters

      class SegmentAssignment(val segmentCenter: Position) extends Ensemble {
        name(s"Assignment for field ${field.idx} @ ${segmentCenter.x},${segmentCenter.y}")
        
        val drone = oneOf(operationalDrones)|\label{inlst:oneOf}|

        utility { drone.sum(x => dist2Utility(x.position, segmentCenter)) }
        tasks { moveTask(drone, segmentCenter) }
      }
      val protectionSegmentAssignments = rules(segmentCenters.map(new SegmentAssignment(_)))|\label{inlst:rules}|
      
      utility { protectionSegmentAssignments.sum(assignment => assignment.utility) / droneCount }
      constraint( protectionSegmentAssignments.map(_.drone).allDisjoint )
    }

    class PatrolUnknown(val field: FieldComponent) extends Ensemble { /* ... */ }

    class ChargerAssignment(charger: ChargerComponent) extends Ensemble { /* ... */ }

    val patrolUnknown = ensembles(fieldsWithUnknownStatus.map(new PatrolUnknown(_))) |\label{inlst:ensembles}|
    val chargerAssignments = ensembles(freeChargers.map(new ChargerAssignment(_)))
    val approachFieldsUnderThreat = ensembles(fieldsUnderThreat.filter(ApproachFieldUnderThreat.isInSituation(_)).map(new ApproachFieldUnderThreat(_)))
    val scareFormations = ensembles(fieldsUnderThreat.filter(ScareFormation.isInSituation(_)).map(new ScareFormation(_)))

    utility { |\label{inlst:utility}|
      approachFieldsUnderThreat.sum(assignment => assignment.utility) +
      scareFormations.sum(assignment => assignment.utility) +
      patrolUnknown.sum(assignment => assignment.utility) / 4 +
      chargerAssignments.sum(assignment => assignment.utility)
    }
    constraint(|\label{inlst:constraint}|
      (patrolUnknown.map(_.drone) ++ approachFieldsUnderThreat.map(_.drones) ++ scareFormations.map(_.drones)).allDisjoint &&
      chargerAssignments.map(_.drone).allDisjoint
    )
  }

  val root = EnsembleSystem(new DroneProtectionSystem)|\label{inlst:root}|
}
\end{lstlisting}

The more complete version of the example is available in \cite{hnetynka_using_2020} together with details of the ensemble definition and DSL definition.
The complete code of the example is available at \url{https://github.com/smartarch/afcens}.

\section{Methods}
\label{sec:methods}
\subsection{As a Constraint Satisfaction Problem}
\label{as-csp}

Our existing approach to instantiating ensembles (described, e.g., in \cite{tomas_bures_language_2020}) is to cast it as a constraint satisfaction (CSP) and optimization problem. That is, how to assign given component instances to ensemble instances such that the cardinality restrictions (e.g., line~\ref{inlst:oneOf} or line~\ref{inlst:subsetOf} in Listing~\ref{lst:dsl}) and constraint blocks in ensembles (e.g., line~\ref{inlst:constraint}) are satisfied and such that the utility function of the root ensemble instance is maximized. 

The ensemble specification (Listing~\ref{lst:dsl}) describes all potential ensemble instances and for each potential ensemble instance, it defines its potential member component instances. However, not all of these potential ensemble instances are eventually created. Only those are created that together satisfy the constraints while maximizing the utility function. Technically, existence of each potential ensemble instance is represented in the constraint optimization problem by a Boolean variable. Similarly, membership of a component instance in a particular ensemble instance is again represented by a Boolean variable. Constraints are formed to reflect the structural constraints between ensemble component instances (such as those that existence of a sub-ensemble instance implies existence of the parent ensemble instance) and the constraints expressed in the specification. The result of the constraint optimization is the assignment of Boolean variables that indicate which ensemble instances are to be created and which components are their members. We perform the constraint optimization using an existing CSP solver (in particular Choco Solver~\footnote{\url{https://choco-solver.org/}}). 

An obvious problem with this approach is that the constraint optimization has by its nature exponential complexity. We performed several test on various scenarios and all indicate that while the approach works well and has clear and crisp semantics, it does not scale beyond a dozen or several dozens of component instances (depending on the complexity of the specification). 
Beyond such limit, the CSP solver requires minutes or even hours to run, which makes it impractical for forming ensembles at runtime as this requires periodic re-evaluation of ensembles (for instance every couple of seconds or minutes). As such, we explore further in this section an alternative way to decide on which ensembles to instantiate by re-casting the ensemble forming as a classification problem. Though the training phase of the classifier takes long, the actual execution of the classifier is almost instantaneous---thus very suitable for constant re-evaluation of the situation at runtime.

\subsection{As a Classification Problem}
\label{as-classification}

When casting the ensemble resolution as a classification problem, we take the inputs and outputs of the CSP solver as a starting point and, instead of invoking a CSP solver to determine the assignment of component instances to ensemble instances, we train classifiers to predict such assignment given the same inputs as the CSP solver. 
Our approach is application of supervised learning where each classifier is trained on a number of examples of inputs and outputs provided by the historical invocations of the CSP solver.

The inputs of the classification problem are identical to the CSP problem inputs and comprise the component knowledge (Listing~\ref{lst:dsl}) that influences in any way the ensemble resolution process, i.e. is considered in the constraints expressed in the specification. 
In our running example, the knowledge fields included are the ones presented in Table~\ref{tbl:inputs}. 

\begin{table}[]
	\centering
	\begin{tabular}{cccc}
		\toprule
		\textbf{Component} & 
		\textbf{Field} &
		\textbf{Domain} & 
		\textbf{Example}
		\\ 
		\midrule
		\texttt{Charger}  
		& \texttt{occupied}  &  
		\texttt{Boolean}  &           
		\texttt{True}
		\\
		\midrule
		\multirow{4}{*}{\begin{tabular}[c]{@{}l@{}}\texttt{Drone}\end{tabular}} 
		& \texttt{energy} &
		\texttt{float} &                                                                 
		0.82
		\\
		& \texttt{x} &                                                                  
		\texttt{float}&                                                                 
		113.125
		\\ 
		& \texttt{y} &                                                                  
		\texttt{float}&                                                                 
		53.375 
		\\
		& \texttt{mode} &                                                                  
		\texttt{enum}&                                                                 
		\texttt{CHARGING}
		\\
		\midrule
		\multirow{2}{*}{\begin{tabular}[c]{@{}l@{}}\texttt{Flock}\end{tabular}} 
		& \texttt{x} &
		\texttt{float} &                                                                 
		157.875
		\\
		& \texttt{y} &                                                                  
		\texttt{float}&                                                                 
		 66.187
		\\ 
		\cmidrule(l){1-4} 
	\end{tabular}
	\caption{Inputs of classification problem.}
	\label{tbl:inputs}
\end{table}

The outputs of the CPS problem are the Boolean variables that represent membership of a component instance to an ensemble instance. 
For the classification problem, we use as outputs nominal variables that represent the membership of component instances to ensembles. 

Since there are several outputs and the outputs follow a nominal scale, the overall classification problem we need to solve to assign component instances to ensemble instances in our setting is a multi-output, multi-class problem. 
Such problems can be solved by training either \textit{one} multi-output, multi-class classifier (which simultaneously predicts all outputs) or \textit{several} independent single-output, multi-class classifiers (one for each output). 
In our experiments, we used both single-output and multi-output classifiers and different learning methods, namely decision trees and neural networks, overviewed next. 

\paragraph{Decision Trees (DT)} represent a supervised learning method that can be used both for classification and regression. 
Its operation is based on creating a tree structure where leaves represent class labels (in case of classification) or continuous values (in regression) and branches represent decision rules inferred from the values of the inputs that lead to the leaves. 
In the case of multi-output problems, leaves represent a set of class labels, one for each output.
A DT is iteratively constructed by an algorithm that selects, at each iteration, a value for an input that splits the output dataset into two subsets and uses a metric to evaluate the homogeneity of the output within the subsets.
In our experiments, we used the Gini impurity metric~\cite{raileanu_theoretical_2004}. We did not use any pruning or set a maximum depth to the derived trees. 
Finally, since we deal with unbalanced data (Fig.~\ref{fig:supports}), we use class weighting in training, with the inverse of the class distribution in our training sets.
We experimented with two DT variants: the \textit{multi-output} variant, in which a single DT was trained to predict the decisions of four drones (we used this number of drones for all our experiments, see Section~\ref{experimental-setup}) and the \textit{single-output} variant, in which four separate DTs were trained, one for each drone. 

\paragraph{Neural networks (NN).}

\begin{figure}[t]
	\includegraphics[width=\columnwidth]{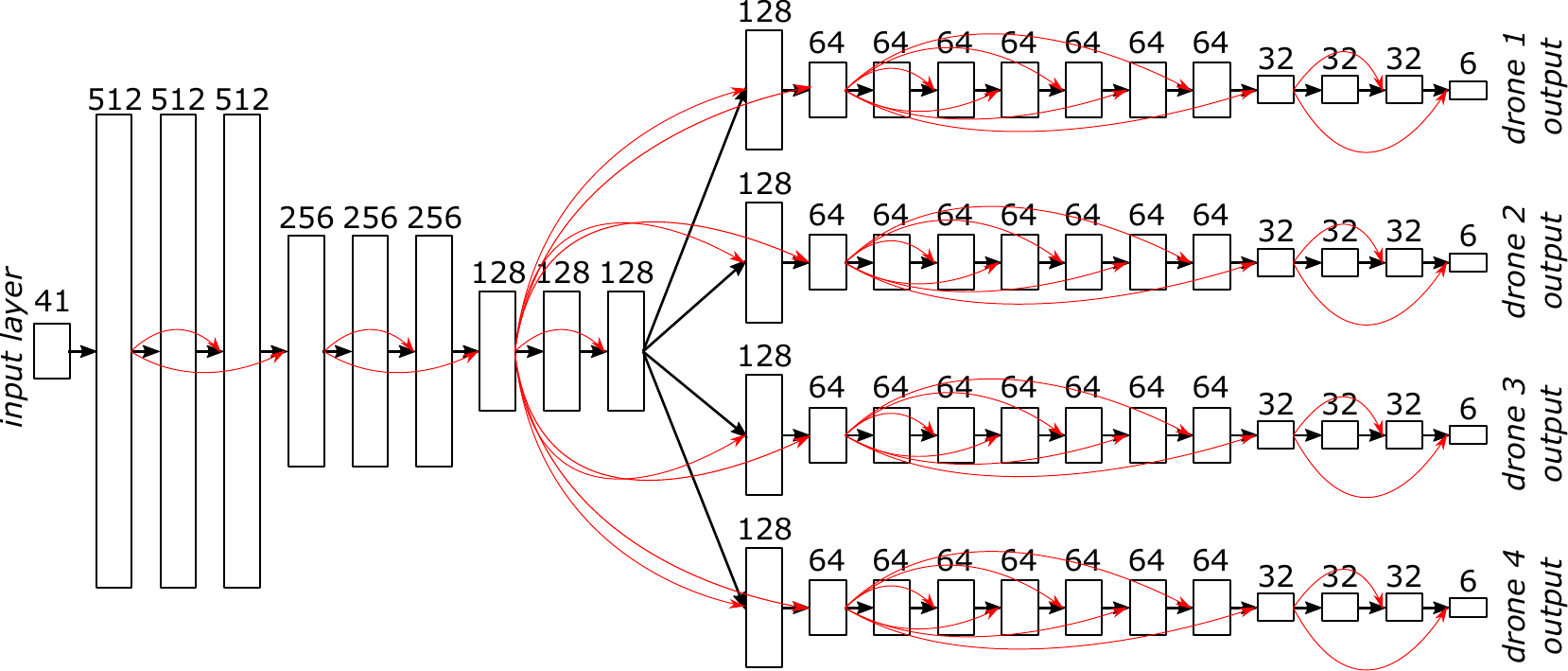}
	\caption{Architecture of \textit{large NN} with emphasized residual connections. Numbers denote width of the fully connected layers.}
	\label{fig:network}
\end{figure}

\begin{figure}[t]

	\centering
	\includegraphics[width=5.5cm]{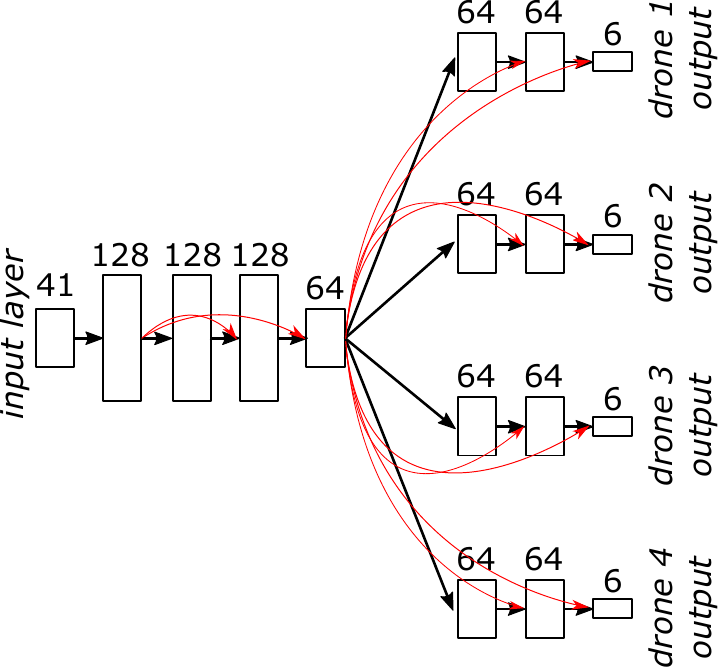}
	\caption{Architecture of \textit{small NN} with emphasized residual connections. Numbers denote width of the fully connected layers.}
	\label{fig:network_small}
\end{figure}

In our experiments, we used fully connected feed-forward NNs with residual connections that connect layers that are not adjacent. In between the dense layers and within the first layer we regularize the outputs with batch normalization~\cite{batchnorm}. 
As a optimizer we use LAMB~\cite{you_lamb_2019}---(stochastic gradient descent method that is based on Layer-wise adaptive estimation of first-order and second-order moments) with multiple steps of logarithmic learning decay. 
We used a batch size of 50.000 and trained the NNs for 50 epochs.
We experimented with different network architectures---Figure~\ref{fig:network} shows the architecture of the best performing NN (called \textit{large NN} henceforth), figure~\ref{fig:network_small} a shows less complex, but only slightly worse performing network, which we term \textit{small NN}. 
Both networks have four outputs corresponding to the decisions of the four drones.

\section{Evaluation}
\label{sec:evaluation}

In this section, we provide an initial evaluation of the feasibility of our prediction-based ensemble resolution approach.
First, we describe the experimental setup used for both obtaining historical data and for evaluating the use of the trained classifiers at runtime. 
Then, we overview the classification performance of the different learning methods using different metrics.  
Finally, we describe the results obtained when using the trained classifiers, together with the CSP solver, for ensemble resolution at runtime. 

\subsection{Experimental Setup}
\label{experimental-setup}

Our experimental setup consists of a simulation that runs the smart farming system described in Section~\ref{sec:example}.
In all our experiments, we have used the following number of component instances: four \texttt{drones}, five \texttt{fields}, three \texttt{chargers}, and five \texttt{flocks}. 
Instead of considering all the component instances and all the potential ensembles in the evaluation of our prediction-based approach, in our preliminary experiments we focused on the \texttt{drone} component and the \texttt{ApproachFieldUnderThreat} ensemble.  
We created classifiers to predict four output variables, each corresponding to one \texttt{drone} in the simulation, capturing whether the \texttt{drone} belongs to the \texttt{ApproachFieldUnderThreat} ensemble formed for the protection of a particular field.
So, each output variable takes one of six values: ``no field'', ``field 1'', ``field 2'', ``field 3'', ``field 4'', ``field 5''.

\subsection{Evaluation of classification performance}

To evaluate the performance of the different classifiers we trained, we used the metrics of balanced accuracy, precision, recall, and F1-score.
These metrics are computed as follows. 
We count true positives (TP), true negatives (TN), false positives (FP), and false negatives (FN) that characterize whether a class has been correctly predicted by a classifier.
In particular, for each of the six classes (``no field'', ``field 1'', ``field 2'', ``field 3'', ``field 4'', ``field 5'') and for each output variable, TP, TN, FP, and FN numbers are calculated in the following way. 
Given a predicted value $y_{pred}$ and a real value $y_{real}$ and considering a class \textit{c}:
\begin{enumerate}[(a)]
	\item $TP_c$ is increased by one if $y_{pred} = c$ and $y_{real} = c$;
	\item $TN_c$ is increased by one if $y_{pred} \neq c$ and $y_{real} \neq c$;
	\item $FP_c$ is increased by one if $y_{pred} = c$ and $y_{real} \neq c$;
	\item $FN_c$ is increased by one if $y_{pred} \neq c$ and $y_{real} = c$.
\end{enumerate}

Given the above, the following metrics can be defined for a class \textit{c}:
\begin{inparaenum}[(i)]
	\item $accuracy_c= \frac {TP_c+TN_c} {TP_c+TN_c+FP_c+FN_c}$, capturing the ratio of correct predictions; 
	\item $precision_c=\frac {TP_c} {TP_c+FP_c}$, capturing the time a positive prediction was also correct; 	
	\item $recall_c=\frac {TP_c} {TP_c+FN_c}$, capturing the times a positive value was also predicted as such; and 
	\item $F1 score_c=\frac {2* precision_c * recall_c} {precision_c + recall_c}$, the harmonic mean of precision and recall.
\end{inparaenum}

\begin{figure}[t]
	\centering
	\includegraphics[width=6cm]{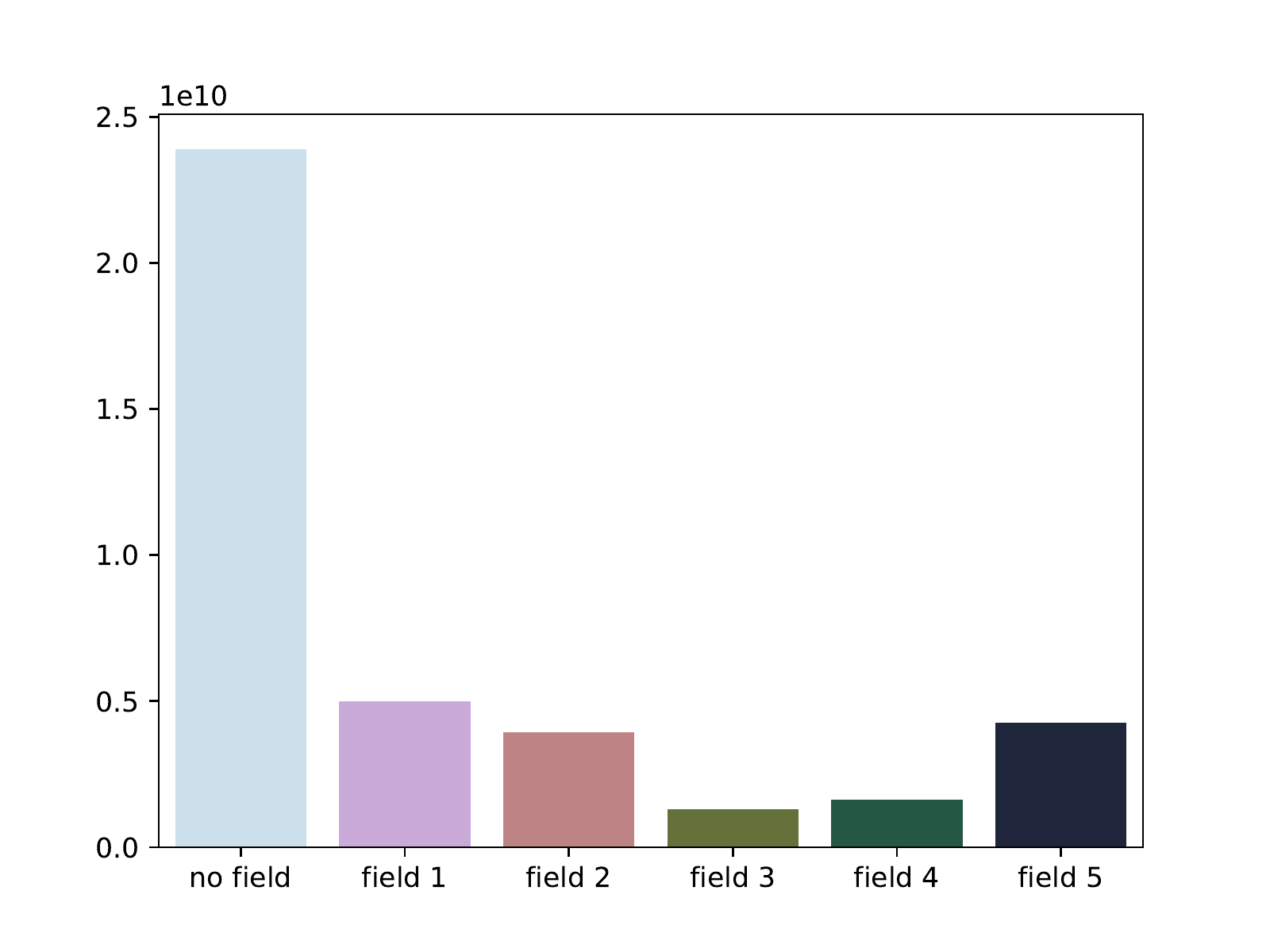}
	\caption{Distribution of dataset in different classes. }
	\label{fig:supports}
\end{figure}

To give a single indicator of classification performance, the above metrics can be averaged over the different six classes in our dataset.
A popular overall metric in multi-class classification is the average accuracy calculated by simply averaging the accuracy from all classes. 
However, such calculation can lead to biased estimates of classifier performance when tested on imbalanced datasets. 
In our case, we indeed have a imbalanced datasets, since the ``no ensemble'' class appears much more often than the other classes (Fig.~\ref{fig:supports}).
We have thus opted for calculating the \textit{balanced accuracy} by first dividing each class accuracy by the number of instances of that class and then taking the average~\cite{brodersen_balanced_2010}.

\begin{table}
	\begin{tabular*}{\textwidth}{l @{\extracolsep{\fill}} ccc}		
		\toprule
		Training samples (million) & 
		1 & 
    	10 &
		100
		\\ 
		\midrule
		multi-output decision tree &
		49.11\% &
		62.29\% &
		72.33\%
		\\
		single-output decision trees &
		70.18\% &
		77.21\% &
		83.01\%
		\\
		small neural network &
		71.87\% &
		94.02\% &
		97.03\%
		\\
		large neural network &
		75.55\% &
		95.86\% &
		98.77\%
		\\
		\bottomrule
		\\
	\end{tabular*}
	\caption{Average balanced accuracy of the different classifiers, calculated with testing set of 100 million samples, for three sizes of training sets.}
	\label{tbl:balanced-accuracies}
\end{table}

Table~\ref{tbl:balanced-accuracies} depicts the balanced accuracy of the different classifiers we have trained, averaged over the four outputs.
The testing set was always set to 100 million samples. 
A first observation is that the four separate single-output decision tree classifiers outperform the multi-output decision tree one (which performed overall very poorly). 
Another observation is that the two neural networks outperform the decision tree classifiers in all cases, while big improvements are seen compared to the decision trees when trained with 10 and 100 million samples. 
Finally, for all classifiers, there is always an improvement when trained with more data; however, while for decision trees the improvement is linear (e.g. $\sim$7\% from one to 10 million and $\sim$6\% from 10 to 100 million for the single-output case), for the neural networks it is more profound when going from one to 10 million ($\sim$20-23\%) compared to going from 10 to 100 million ($\sim$3\%).

\begin{figure}[t]
	\centering
	\includegraphics[width=\linewidth]{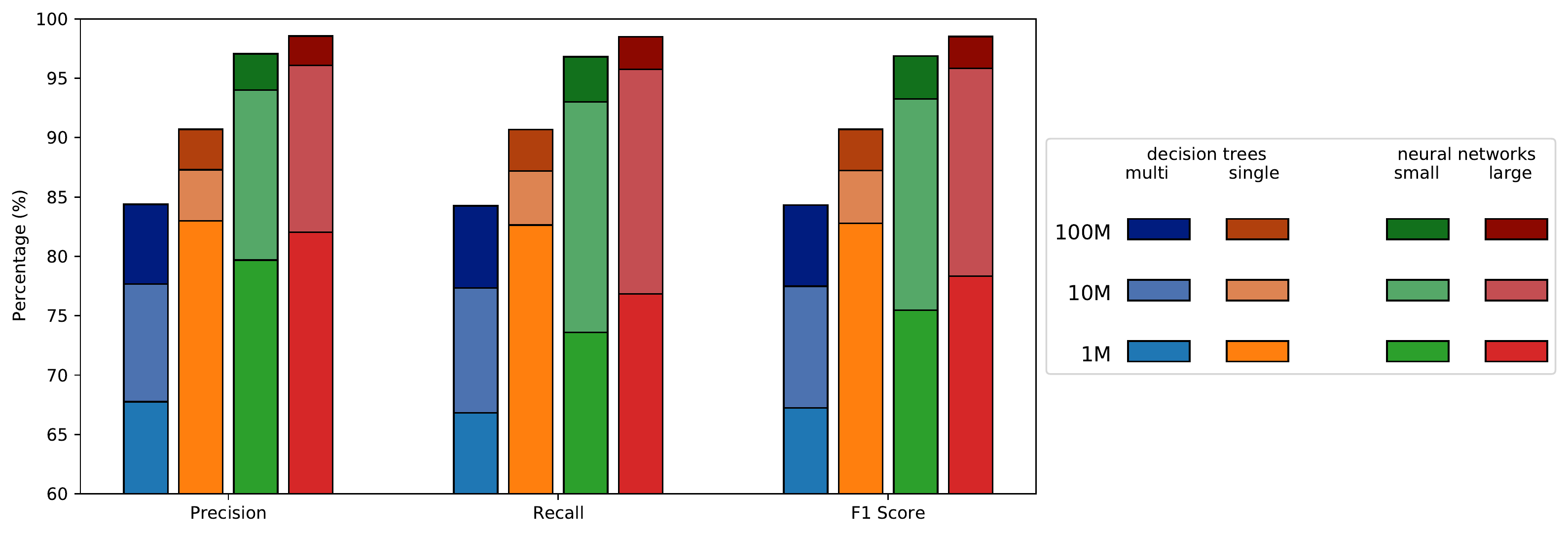}
	\caption{Weighted averages of precisions, recalls, and F1-scores for different classifiers and training set sizes, in million (10M depicts the delta between the result obtained with 10M and 1M samples and 100M the delta between 100M and 10 M samples).}
\label{fig:results-scores}
\end{figure}

We also calculated the weighted average of precision, recall, and F1 score for each drone. 
The weighted average is calculated by weighting each class according to its inverse prevalence (Figure~\ref{fig:supports}). 
Figure~\ref{fig:results-scores} shows the weighted averages of the three metrics, averaged over the four drones. 
We observe that the single-output classifiers perform universally better than the multi-output one; similarly, the large neural network performs slightly better than the small one across the board. 
Also, we see the same pattern as with weighted averages: in neural networks, there is a large gain when using 10 instead of 1 million data; this gain is much smaller when using 100 instead of 10 million data. 
Instead, for decision trees, there is an almost equal gain when using 10 times more data.  
Finally, contrary to the balanced accuracy results, the single-output decision tree classifier outperforms the other classifiers (including the neural network ones) in all metrics when trained with 1 million data.

\subsection{Experiments using Classifiers for Ensemble Resolution}

To evaluate how effective the trained classifiers are when used together with the CSP solver in ensemble resolution, we plugged them in the simulation of our running example and evaluated the overall utility of a run, given by the amount of undisturbed time that birds spend on fields. 
We compared the following cases: 
\begin{itemize}
	\item Runs where the drones were never participating in the \texttt{ApproachFieldUnder\-Threat} ensemble. This serves as the overall baseline.
	\item Runs with ensemble resolution every 1 minute of simulation time by invoking the CSP solver. 
	\item Runs with ensemble resolution every 1 minute of simulation time by means of invoking a respective predictor to determine the participation of each \texttt{drone} in \texttt{ApproachField\-UnderThreat}. (As the other ensembles were not handled by the predictor, we still formed them based on the CSP solver.)
\end{itemize}	

\begin{figure}[t]
	\centering
	\includegraphics[width=\textwidth]{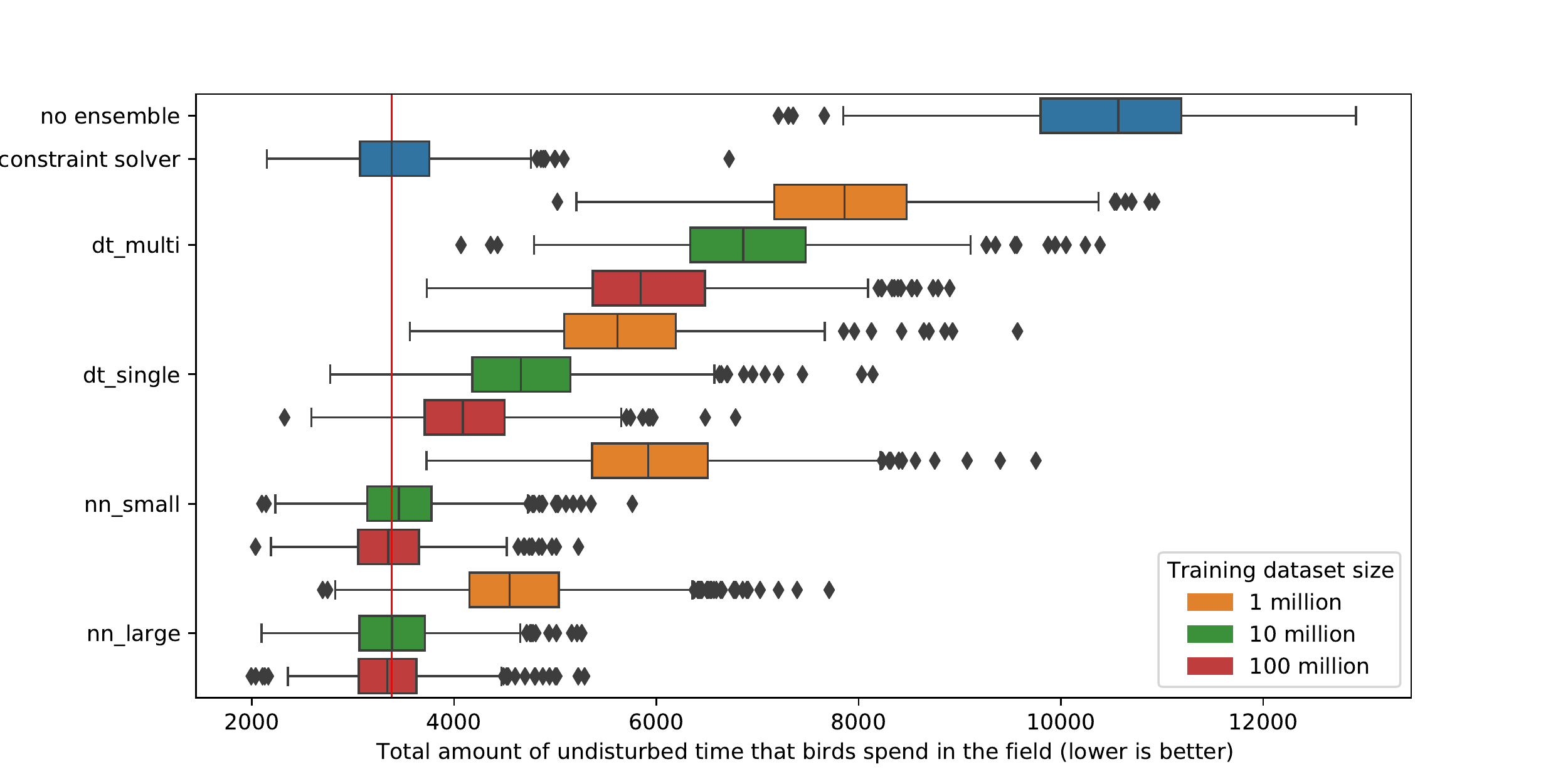}
	\caption{Boxplots of overall utilities (undisturbed time for birds) when plugging the different predictors in the simulation of our running example.}
	\label{fig:boxplot}
\end{figure}

Each run had a duration of 600 minutes of simulated time; we performed 1000 runs (randomizing the starting position of flocks and drones) for each case, with the results depicted in Fig.~\ref{fig:boxplot}.

The boxplots show that, for each predictor, using more training data resulted in higher overall utility. 
Also, while in most of the cases, using the predictor with the CSP resulted in a reduction of the system utility w.r.t using only the CSP solver, all cases performed better than not participating in ensembles at all. 
Also, in two cases (\texttt{nn\_small} and \texttt{nn\_large} trained with 100 million samples) there was a marginal improvement over the CSP case. 
We believe this happens due to the classifier being able to generalize over the solution provided in the form of hard/soft constraints and was able to improve on corner cases which lead for instance to short intervals of oscillation when a drone was switching between two targets and, as the result, did not move.
Another interesting observation is that higher balanced accuracy does not always translates to higher utility in the simulation: e.g. even though the \texttt{small\_nn} on 1 million samples had slightly higher balanced accuracy than the respective \texttt{dd\_single} (Table~\ref{tbl:balanced-accuracies}), its overall utility is slightly lower, as can be seen in Fig.~\ref{fig:boxplot}.

\section{Related work}
\label{sec:related-work}

The original idea of ensembles is based on the SCEL language~\cite{nicola_formal_2014}.
In its implementation JRESP\footnote{\url{http://jresp.sourceforge.net/}}~\cite{nicola_formal_2014}, the ensembles are formed implicitly, as they are abstractions capturing groups of components and dynamically determined by their attribute-based communication.
Almost the same approach is used in the $Ab^aCuS$~\cite{alrahman_programming_2016} (which is not strictly an ensemble-based framework however it is built on the principles of the attribute-based communication).
Another ensemble-based framework is Helena~\cite{hennicker_foundations_2014} but here, the ensembles are formed explicitly, i.e., the components indicate, to which the ensemble they belong.
In our implementation~\cite{tomas_bures_language_2020} of ensemble-based system, we have used the constraint solver for forming ensembles but its issues have been already mentioned in Section~\ref{sec:methods}.

The concept of ensembles targets emergent configurations.
In the implementation of the ECo-IoT~\cite{alkhabbas_eco-iot_2018} approach, the emergent configurations are solved using a state space planner, nevertheless its performance is also hindered by exponential complexity.
The planner is also employed in \cite{bucchiarone_collective_2019}, where the linear complexity is claimed. 
However, measurements are done with respect to the growing number of adaptations in a system (and not the size of the system).

Machine learning has been recently used in different ways to replace heavy computations that are part of constraint satisfaction and combinatorial optimization problems by fast approximations using machine learning, as overviewed in a recent survey~\cite{bengio_machine_2020}. 
For example, decision trees have been used in predicting the satisfiability of SAT instances~\cite{xu_predicting_2012}, while deep neural networks have been used in predicting the satisfiabilities of random Boolean binary CSPs with high prediction accuracies (>99.99\%)~\cite{hooker_towards_2018}.
Other approaches have tried to embed neural networks, but also decision trees and random forests, in constraint programming~\cite{lee_neuron_2011,michel_embedding_2015}. 
The idea is to learn part of the combinatorial optimization model and embed the learned piece of knowledge in the combinatorial model itself. 
We too train classifiers to overcome computational issues associated with ensemble resolution, where constraint programming is the host technology.

\section{Conclusion}
\label{sec:conclusion}

In this paper, we proposed a new approach for formation of ensembles at runtime. Instead of relying on a constraint solver to decide on optimal placement of components to ensembles, we employed machine learning. This allows us to tackle the problem of exponential time needed by the constraint solver, which is especially a problem since ensembles have to be formed at runtime (typically in real-time). When employing machine learning, we can (after the initial training stage) decide on placement of components into ensembles with linear time.

In our approach, we casted the problem of ensemble formation to a classification problem. To give comparison how well this approach works, we implemented two different classifiers---decision trees and neural networks. We show that with enough data, we were able to train the predictor with high enough accuracy. Not only that, when we plugged the predictor in the simulation of our running example and evaluated the overall utility, we observed that some classifiers even perform marginally better than the original solution that employed the constraint solver. We attribute this to the fact that the classifier was able to generalize the solution provided in the form of ensemble specification (i.e., logical predicates and optimization function).

This paper provided the initial idea and indicative experiments. In future work, we would like to focus on generalization which would allow the predictors to be trained on smaller models and generalize them to larger scope. Also, we are looking into how to further improve the prediction and how to include reinforcement learning to take also into account the overall end-to-end utility function (e.g. the time that the birds spend undisturbed in the field -- as used in our running example).

\section*{Acknowledgment}

This paper is part of a project that has received funding from the European Research Council (ERC) under the European Union’s Horizon 2020 research and innovation programme (grant agreement No 810115). Also, the research leading to these results has received funding from the ECSEL Joint Undertaking (JU) under grant agreement No 783221 and was partially supported by Charles University institutional funding SVV 260451.

We are also grateful to Milan Straka from Institute of Formal and Applied Linguistics at Faculty of Mathematics and Physics at Charles University for valuable input in the field of deep networks that improved the training speed and results significantly.

\bibliographystyle{splncs04}
\bibliography{paper}

\begin{thebibliography}{10}
\providecommand{\url}[1]{\texttt{#1}}
\providecommand{\urlprefix}{URL }
\providecommand{\doi}[1]{https://doi.org/#1}

\bibitem{alkhabbas_eco-iot_2018}
Alkhabbas, F., Spalazzese, R., Davidsson, P.: {ECo}-{IoT}: {An} {Architectural}
  {Approach} for {Realizing} {Emergent} {Configurations} in the {Internet} of
  {Things}. In: Proceedings of {ECSA} 2018, {Madrid}, {Spain}. {LNCS}, vol.
  11048, pp. 86--102. Springer (Sep 2018). \doi{10.1007/978-3-030-00761-4\_6}

\bibitem{alrahman_programming_2016}
Alrahman, Y.A., Nicola, R.D., Loreti, M.: Programming of {CAS} {Systems} by
  {Relying} on {Attribute}-{Based} {Communication}. In: Proceedings of {ISOLA}
  2016, {Corfu}, {Greece}. {LNCS}, vol.~9952, pp. 539--553. Springer (Oct
  2016). \doi{10.1007/978-3-319-47166-2\_38}

\bibitem{lee_neuron_2011}
Bartolini, A., Lombardi, M., Milano, M., Benini, L.: Neuron {Constraints} to
  {Model} {Complex} {Real}-{World} {Problems}. In: Lee, J. (ed.) Principles and
  {Practice} of {Constraint} {Programming} – {CP} 2011, vol.~6876, pp.
  115--129. Springer Berlin Heidelberg (2011).
  \doi{10.1007/978-3-642-23786-7\_11}

\bibitem{bengio_machine_2020}
Bengio, Y., Lodi, A., Prouvost, A.: Machine {Learning} for {Combinatorial}
  {Optimization}: a {Methodological} {Tour} d'{Horizon}. arXiv:1811.06128 [cs,
  stat]  (Mar 2020), \url{http://arxiv.org/abs/1811.06128}, arXiv: 1811.06128

\bibitem{michel_embedding_2015}
Bonfietti, A., Lombardi, M., Milano, M.: Embedding {Decision} {Trees} and
  {Random} {Forests} in {Constraint} {Programming}. In: Michel, L. (ed.)
  Integration of {AI} and {OR} {Techniques} in {Constraint} {Programming},
  vol.~9075, pp. 74--90. Springer International Publishing, Cham (2015).
  \doi{10.1007/978-3-319-18008-3\_6}

\bibitem{brodersen_balanced_2010}
Brodersen, K.H., Ong, C.S., Stephan, K.E., Buhmann, J.M.: The {Balanced}
  {Accuracy} and {Its} {Posterior} {Distribution}. In: Proceedings of ICPR
  2010, Istanbul, Turkey. pp. 3121--3124. IEEE (Aug 2010).
  \doi{10.1109/ICPR.2010.764}

\bibitem{bucchiarone_collective_2019}
Bucchiarone, A.: Collective {Adaptation} through {Multi}-{Agents} {Ensembles}:
  {The} {Case} of {Smart} {Urban} {Mobility}. ACM Transactions on Autonomous
  and Adaptive Systems  \textbf{14}(2),  1--28 (Dec 2019).
  \doi{10.1145/3355562}

\bibitem{hennicker_foundations_2014}
Hennicker, R., Klarl, A.: Foundations for {Ensemble} {Modeling} – {The}
  {Helena} {Approach}. In: Specification, {Algebra}, and {Software}, pp.
  359--381. No.~8373 in {LNCS}, Springer (2014).
  \doi{10.1007/978-3-642-54624-2\_1}

\bibitem{hnetynka_using_2020}
Hnetynka, P., Bures, T., Gerostathopoulos, I., Pacovsky, J.: Using {Component}
  {Ensembles} for {Modeling} {Autonomic} {Component} {Collaboration} in {Smart}
  {Farming}. In: Proceedings of {SEAMS} 2020 ({Accepted}), {Seoul}, {Korea}
  (2020)

\bibitem{batchnorm}
Ioffe, S., Szegedy, C.: Batch normalization: Accelerating deep network training
  by reducing internal covariate shift. arXiv:1502.03167 [cs]  (Mar 2015),
  \url{http://arxiv.org/abs/1502.03167}, arXiv: 1502.03167

\bibitem{nicola_formal_2014}
Nicola, R.D., Loreti, M., Pugliese, R., Tiezzi, F.: A {Formal} {Approach} to
  {Autonomic} {Systems} {Programming}: {The} {SCEL} {Language}. ACM
  Transactions on Autonomous and Adaptive Systems  \textbf{9}(2),  7:1--7:29
  (Jul 2014). \doi{10.1145/2619998}

\bibitem{raileanu_theoretical_2004}
Raileanu, L.E., Stoffel, K.: Theoretical {Comparison} between the {Gini}
  {Index} and {Information} {Gain} {Criteria}. Annals of Mathematics and
  Artificial Intelligence  \textbf{41}(1),  77--93 (May 2004).
  \doi{10.1023/B:AMAI.0000018580.96245.c6}

\bibitem{tomas_bures_language_2020}
{Tomas Bures}, {Ilias Gerostathopoulos}, {Petr Hnetynka}, {Frantisek Plasil},
  {Filip Krijt}, {Jiri Vinarek}, {Jan Kofron}: A {Language} and {Framework} for
  {Dynamic} {Component} {Ensembles} in {Smart} {Systems}. Int. Journal on
  Software Tools for Technology Transfer (STTT))  (2020).
  \doi{https://doi.org/10.1007/s10009-020-00558-z}

\bibitem{hooker_towards_2018}
Xu, H., Koenig, S., Kumar, T.K.S.: Towards {Effective} {Deep} {Learning} for
  {Constraint} {Satisfaction} {Problems}. In: Hooker, J. (ed.) Principles and
  {Practice} of {Constraint} {Programming}, pp. 588--597. Springer, Cham
  (2018). \doi{10.1007/978-3-319-98334-9\_38}

\bibitem{xu_predicting_2012}
Xu, L., Hoos, H.H., Leyton-Brown, K.: Predicting satisfiability at the phase
  transition. In: Proceedings of the {Twenty}-{Sixth} {AAAI} {Conference} on
  {Artificial} {Intelligence}. pp. 584--590. {AAAI}'12, AAAI Press, Toronto,
  Ontario, Canada (Jul 2012)

\bibitem{you_lamb_2019}
You, Y., Li, J., Reddi, S., Hseu, J., Kumar, S., Bhojanapalli, S., Song, X.,
  Demmel, J., Keutzer, K., Hsieh, C.J.: Large batch optimization for deep
  learning: Training {BERT} in 76 minutes. arXiv:1904.00962 [cs.LG]  (2019),
  \url{https://arxiv.org/abs/1904.00962}

\end{thebibliography}

\end{document}